%% file: acl_latex.tex
\documentclass[11pt]{article}

\usepackage[preprint]{acl}

\usepackage{times}
\usepackage{latexsym}

\usepackage[T1]{fontenc}
\usepackage[utf8]{inputenc}

\usepackage{microtype}

\usepackage{inconsolata}

\usepackage{graphicx}
\usepackage[ruled,vlined]{algorithm2e}

\usepackage{booktabs}
\usepackage{threeparttable}
\usepackage{array}
\usepackage{multirow}

\usepackage{siunitx}
\usepackage{graphicx}

\title{VeriAgent: A Tool-Integrated Multi-Agent System with Evolving Memory for PPA-Aware RTL Code Generation}

\author{ 
Yaoxiang Wang$^{1}$\;,
Qi Shi$^{2}$\thanks{Corresponding author.},
Shangzhan Li$^{3}$\;,
Qingguo Hu$^{1}$,
Xinyu Yin$^{1}$,
Bo Guo$^{2}$,\\
\textbf{Xu Han$^{2}$, Maosong Sun$^{2}$, Jinsong Su$^{1}$\footnotemark[1]} \\ 
$^{1}$Xiamen University \quad $^{2}$Tsinghua University \quad $^{3}$Harbin Institute of Technology \\ 
}
\begin{document}
\maketitle
\begin{abstract}

LLMs have recently demonstrated strong capabilities in automatic RTL code generation, achieving high syntactic and functional correctness. However, most methods focus on functional correctness while overlooking critical physical design objectives, including Power, Performance, and Area.
% Furthermore, current systems rarely establish a sustained optimization loop with EDA tools, and lack mechanisms to accumulate optimization experience across iterative design cycles.
In this work, we propose a PPA-aware, tool-integrated multi-agent framework for high-quality verilog code generation. Our framework explicitly incorporates EDA tools into a closed-loop workflow composed of a \textit{Programmer Agent}, a \textit{Correctness Agent}, and a \textit{PPA Agent}, enabling joint optimization of functional correctness and physical metrics. To support continuous improvement without model retraining, we introduce an \textit{Evolved Memory Mechanism} that externalizes optimization experience into structured memory nodes. A dedicated memory manager dynamically maintains the memory pool and allows the system to refine strategies based on historical execution trajectories.
Extensive experiments demonstrate that our approach achieves strong functional correctness while delivering significant improvements in PPA metrics. By integrating tool-driven feedback with structured and evolvable memory, our framework transforms RTL generation from one-shot reasoning into a continual, feedback-driven optimization process, providing a scalable pathway for deploying LLMs in real-world hardware design flows.

\end{abstract}

\input{section/1-intro}
\input{section/3-method}
\input{section/4-exp}
\input{section/2-related}

\section{Conclusion}

In this work, we present a PPA-aware tool-integrated multi-agent framework for automated Verilog generation. By coordinating a Programmer Agent, Correctness Agent, and PPA Agent within a closed-loop workflow, the proposed system jointly optimizes functional correctness and physical design metrics through direct interaction with EDA tools. Furthermore, we introduce an evolved structured memory mechanism that continuously accumulates and refines optimization experience across iterations, enabling sustained improvement beyond one-shot generation. Experimental results on VerilogEval and RTLLM demonstrate that our approach maintains strong functional correctness while achieving notable gains in PPA performance. These results highlight the potential of combining tool integration and memory evolution to enable practical LLM-driven hardware design automation.

% \section*{Limitations}

% This document does not cover the content requirements for ACL or any
% other specific venue.  Check the author instructions for
% information on
% maximum page lengths, the required ``Limitations'' section,
% and so on.


\bibliography{custom}

% \appendix

% \section{Example Appendix}
% \label{sec:appendix}

% This is an appendix.

\end{document}

%% file: section/1-intro.tex
\section{Introduction}

With the continuous growth in scale and complexity of integrated circuits, modern chip design flows are facing unprecedented challenges~\citep{chang2023chipgpt}. As a result, manually writing and optimizing Register-Transfer Level (RTL) code has become increasingly costly and time-consuming. In this context, automated RTL code generation~\citep{liu2024rtlcoder} has emerged as a critical direction for shortening design cycles, reducing engineering effort, and improving overall design quality. As the dominant hardware description language, verilog plays a central role in digital circuit design automation. Recent advances in Large Language Models~\citep{achiam2023gpt,team2023gemini, guo2025deepseek,yang2025qwen3} have demonstrated remarkable capabilities in code generation, and have been successfully applied to RTL code generation tasks, achieving significant improvements in syntactic and functional correctness.

Despite these advances, a substantial gap remains before LLM-based RTL code generation can be deployed in industrial settings~\citep{yang2025large}. First, most existing approaches focus primarily on functional correctness, while overlooking physical design metrics that are crucial in real-world workflows, such as Power, Performance, and Area (PPA)~\citep{thorat2025llmveri}. Second, although RTL design inherently relies on Electronic Design Automation (EDA) toolchains such as simulation and synthesis, current methods~\cite{zhao2025mage,tasnia2025veriopt} only leverage environmental feedback in a shallow manner and fail to establish a sustained optimization loop. More importantly, existing systems generally lack a long-term optimization memory mechanism. They are unable to accumulate and refine knowledge across multiple design iterations, making the optimization process heavily dependent on one-shot reasoning rather than continuous improvement.

Existing LLM-based RTL code generation approaches can be broadly categorized into two groups: training-based methods and training-free methods. Training-based methods~\citep{pei2024betterv, liu2024rtlcoder,zhang2025qimeng,zhao2025codev} typically rely on large-scale synthetic data for supervised fine-tuning or reinforcement learning to enhance model performance on RTL tasks. However, these methods often depend on automatically generated samples whose distributions exhibit significant bias (e.g., simple modules and templated patterns), limiting their ability to generalize to complex design scenarios. Moreover, training incurs substantial computational cost, and in practice, the resulting models often underperform compared to directly using state-of-the-art closed-source commercial LLMs. 

Training-free methods aim to improve generation quality without modifying model parameters. Multi-agent frameworks have been introduced to decompose RTL code generation into collaborative stages. For example, ~\citet{zhao2024magemultiagentengineautomated} and ~\citet{wu2024chateda} improve functional correctness by organizing planning, generation, and verification into coordinated phases. However, these systems primarily focus on syntax and functional validation, lacking systematic modeling of PPA-related signals. The rich timing, area, and power information produced by EDA tools is not abstracted into reusable knowledge, but instead remains as single-round feedback. Some works incorporate PPA constraints into verilog generation. For instance, ~\citet{tasnia2025veriopt} and ~\citet{thorat2025llmveri} leverage in-context learning for PPA-aware optimization, yet its effectiveness largely depends on prompt engineering and lacks an automated experience accumulation mechanism.

To address these limitations, we propose a tool-integrated multi-agent framework for generating high-quality RTL code, equipped with an evolvable structured memory mechanism to enable continuous improvement. First, we design a PPA-aware Tool-Integrated Multi-Agent workflow that explicitly incorporates EDA tools into a closed-loop generation process. The system consists of three cooperative agents: a \textit{Programmer Agent} responsible for code generation, a \textit{Correctness Agent} that executes testbenches, and a \textit{PPA Agent} that invokes synthesis tools and analyzes PPA metrics. This design enables joint optimization of functional correctness and physical design objectives. 
Furthermore, we introduce an \textit{Evolved Memory Mechanism} to accumulate and refine optimization experience across iterative design cycles. We categorize memory into three complementary types: \textit{Rule Memory}, which captures general design constraints and domain best practices; \textit{Structure Memory}, which abstracts reusable optimization patterns from code structures; and \textit{EDA Signal Memory}, which encodes optimization knowledge derived from PPA evaluation process.
To systematically manage these experiences, we design a dedicated memory manager that extracts memory nodes from the execution trajectories of the multi-agent workflow. The memory manager maintains the memory pool through \textit{discard}, \textit{insert}, and \textit{update} operations, allowing the system to preserve effective strategies, incorporate newly discovered insights, and refine suboptimal guidance. 
Each memory node is structurally defined with explicit triggering conditions and actionable guidance, enabling precise activation during subsequent iterations. By externalizing and evolving optimization experience in this structured manner, our framework transforms RTL code generation from isolated one-shot reasoning into a continual, feedback-driven optimization process.

We conduct comprehensive evaluations on the VerilogEval and RTLLM benchmarks. Experimental results show that our approach maintains high functional correctness while achieving significant improvements in PPA metrics, validating the effectiveness of tool integration and memory evolution. Overall, this work presents a new paradigm for PPA-aware RTL code generation with sustained optimization capability, offering a practical pathway for integrating LLMs into real-world hardware design workflows.

Our contributions are summarized as follows:
\begin{itemize}
    \item We propose a PPA-aware tool-integrated multi-agent framework for automated Verilog generation, which establishes a closed-loop optimization process by coordinating a Programmer Agent, Correctness Agent, and PPA Agent with EDA toolchains to jointly optimize functional correctness and physical design metrics.
    
    \item We introduce an evolved structured memory mechanism that accumulates and refines optimization experience across iterations, consisting of Rule Memory, Structure Memory, and EDA Signal Memory, enabling the system to transform one-shot RTL code generation into a continual feedback-driven optimization process.
    
    \item We conduct extensive experiments on the VerilogEval and RTLLM benchmarks, demonstrating that our method preserves strong functional correctness while achieving significant improvements in PPA metrics, validating the effectiveness of tool-integrated optimization and memory evolution.
\end{itemize}

%% file: section/3-method.tex
\begin{figure*}[ht]
    \centering
    \includegraphics[width=\linewidth]{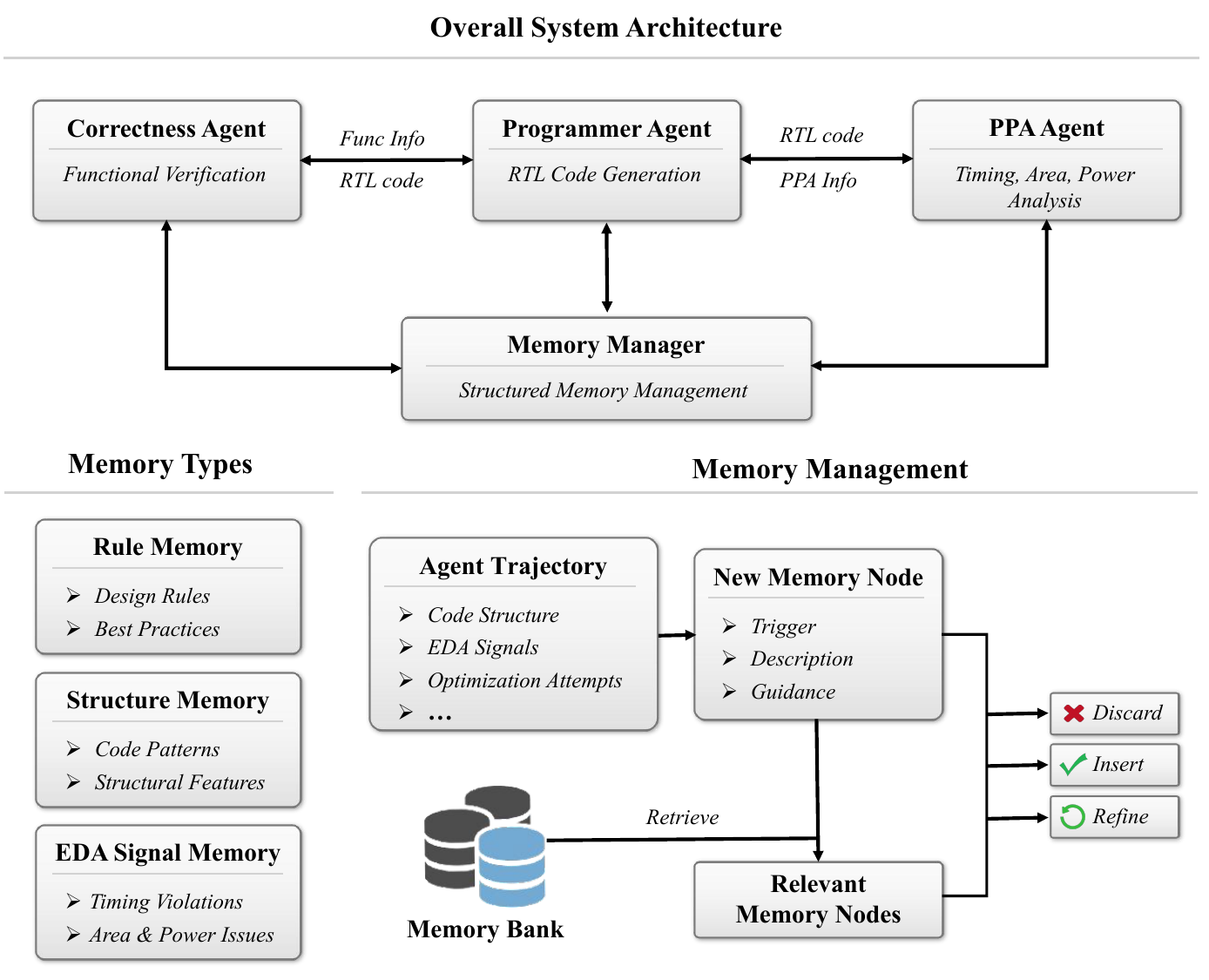}
    \caption{Overview of the proposed PPA-aware multi-agent RTL code generation framework.}
    \label{fig:overview}
\end{figure*}

\section{Methodology}
% In this section, we present our PPA-aware tool-integrated multi-agent framework with an evolved structured memory mechanism. We first introduce the overall system architecture, followed by the multi-agent collaboration workflow, the integration of EDA tools, and finally the design, triggering, and evolution of structured memory.

\subsection{Overall Framework}

As shown in Figure~\ref{fig:overview}, our system consists of four tightly coupled components: a Programmer Agent, a Correctness Agent, a PPA Agent, and a Memory Manager. The Programmer Agent generates verilog implementations according to the design specification. The Correctness Agent performs functional verification through simulation and testbench execution. Once functional correctness is satisfied, the PPA Agent invokes synthesis tools to evaluate timing, area, and power metrics. The Memory Manager monitors the entire execution trajectory, extracts reusable optimization experience, and maintains a structured memory pool.

These components form a closed-loop workflow. In each iteration, the Programmer Agent generates or refines verilog code under the guidance of activated memory. The feedback from both verification and PPA analysis is then abstracted and stored as structured experience, which influences subsequent iterations. Through this mechanism, the system gradually improves both correctness and physical quality without modifying model parameters.

\subsection{Multi-Agent Collaboration}

\paragraph{Programmer Agent.}
The Programmer Agent is responsible for verilog code generation. Unlike previous single-pass prompting approaches, our Programmer Agent conditions its generation on structured memory guidance extracted from previous iterations. This allows the agent to incorporate accumulated optimization knowledge, such as structural refinement strategies or PPA-oriented design patterns, into the current generation step. As a result, code generation becomes progressively informed by historical optimization experience.

\paragraph{Correctness Agent.}
The Correctness Agent ensures functional validity by executing testbenches and analyzing simulation results. The agent abstracts key information from raw error logs, which can be summarized into structured semantic signals, such as interface mismatches, state transition inconsistencies, or arithmetic logic errors. These structured signals not only guide immediate debugging but also serve as inputs to the memory extraction process, enabling the system to generalize recurring functional issues into reusable knowledge.

\paragraph{PPA Agent.}
The PPA Agent integrates synthesis tools into the optimization loop. 
After functional verification succeeds, the agent performs synthesis and extracts key metrics from the logs and reports, including critical path delay, cell areas, and estimated power consumption. The reports can be summarized into higher-level semantic signals such as excessive combinational depth and abnormal area growth. This semantic abstraction bridges low-level tool reports and high-level reasoning, enabling systematic modeling of physical feedback.

\subsection{Evolving Structured Memory}
\subsubsection{Memory Structure}
A key contribution of our framework is the introduction of an evolved structured memory mechanism that externalizes optimization knowledge from iterative execution trajectories. Each memory unit is represented as a structured memory node consisting of trigger conditions, guidance, and metadata. The trigger specifies the semantic conditions under which the memory becomes relevant, which may originate from code structure or EDA feedback. The guidance encodes actionable optimization strategies in natural language. Metadata records the optimization intent (e.g., timing, area, power, correctness) and the applicable agent roles. We categorize memory into three complementary types.

\begin{itemize}
\item{Rule Memory.}
Rule Memory captures general design constraints and domain best practices that are broadly applicable across tasks. These rules encode stable design principles, such as avoiding excessively deep combinational paths or ensuring proper reset handling. Rule Memory provides global guidance and remains active under wide conditions.

\item{Structure Memory.}
Structure Memory is triggered by structural characteristics extracted from RTL code. Through static analysis or pattern detection, the system identifies features such as wide datapaths, deeply nested conditional logic, or replicated arithmetic structures. These structural patterns are associated with corresponding optimization strategies, enabling the system to learn reusable mappings between code forms and improvement actions.

\item{EDA Signal Memory.}
EDA Signal Memory is triggered by semantic signals derived from synthesis and timing analysis. For example, negative slack may activate pipelining strategies, while excessive area may trigger resource sharing or logic refactoring suggestions. By encoding such relationships, the framework transforms transient tool feedback into persistent optimization knowledge.
\end{itemize}

\subsubsection{Memory Triggering}
Before each generation step, the Memory Manager determines which memory nodes should be activated based on the current task context. 
The context consists of the design specification, the current RTL code, and the latest execution feedback.
The Memory Manager performs semantic matching between this context and the trigger fields of memory nodes stored in the memory pool. 
Specifically, the context description together with all candidate triggers are provided to the LLM, which acts as a semantic selector to identify the most relevant triggers and memory nodes. The selected memory nodes are then activated, and their guidance fields are injected into the prompt of the agents to guide the next generation step.

\subsubsection{Memory Evolution}

Beyond memory activation during generation, the Memory Manager continuously evolves the memory pool after each task execution. The goal of this process is to transform the execution trajectory into reusable optimization knowledge and integrate it into the structured memory bank.

Specifically, the execution trajectory $T$ contains the complete interaction history of the current task, including the design specification, generated RTL code, modification steps, simulation feedback, and PPA evaluation results. Based on this trajectory, the Memory Manager extracts potential optimization insights and converts them into candidate memory nodes. Each candidate node contains a trigger condition, optimization guidance, and associated metadata.

The detailed evolution procedure is summarized in Algorithm~1. First, the manager analyzes the trajectory and generates a set of candidate memory nodes $N$ through the function $\textsc{GenerateNodes}(T)$. For each candidate node $n$, the system retrieves the top-$K$ most relevant memory nodes from the memory bank by prompting LLM given the triggers of new node and existing triggers. These retrieved nodes provide contextual references for determining whether the candidate node represents new knowledge or a refinement of existing experience.

Next, the Memory Manager invokes an LLM-based decision module, $\textsc{Decide}(n, R)$, which evaluates the candidate node $n$ together with the retrieved memory set $R$ and outputs an action decision. The decision includes three possible operations: \textit{insert}, \textit{refine}, or \textit{discard}. If the candidate node captures a previously unseen optimization pattern, it is directly inserted into the memory bank. If the node corresponds to an existing memory entry but provides improved or more precise guidance, the corresponding memory node is updated. Otherwise, the candidate node is considered redundant and discarded.

Through repeated execution cycles, this mechanism gradually accumulates structured optimization experience tailored to RTL code generation and PPA-aware design refinement. By externalizing optimization knowledge outside the model parameters, the system converts RTL code generation from isolated one-shot reasoning into a continual, feedback-driven optimization process.

\begin{algorithm}[t]
\caption{Memory Evolution}
\KwIn{Execution trajectory $T$, memory bank $M$, retrieval size $K$}
\KwOut{Updated memory bank $M$}

$N \leftarrow \textsc{GenerateNodes}(T)$

\ForEach{$n \in N$}{
    $R \leftarrow \textsc{Retrieve}(M, n.trigger, K)$ \tcp*{top-K similar nodes}

    $(act, m\_id, content) \leftarrow \textsc{Decide}(n, R)$ \tcp*{act = insert/refine/discard}

    \uIf{$act =$ insert}{
        $M \leftarrow M \cup \{n\}$
    }
    \uElseIf{$act =$ refine}{
        $M[m\_id] \leftarrow content$
    }
    \Else{
        continue
    }
}

\Return $M$
\end{algorithm}

%% file: section/4-exp.tex
\section{Experiments}

\subsection{Setup}

\paragraph{Benchmarks.}
We evaluate our framework on two representative RTL code generation benchmarks: VerilogEval and RTLLM.

\textbf{VerilogEval} is designed to measure the capability of large language models to generate synthesizable Verilog implementations from natural language specifications. It contains two complementary tracks. The Human track consists of design descriptions written by hardware experts, reflecting realistic and diverse specification styles. The Machine track, in contrast, provides model-generated specifications, which are typically more standardized and less ambiguous.

\textbf{RTLLM} evaluates Verilog generation quality across a curated collection of RTL design tasks with varying complexity. The benchmark includes 50 problems covering combinational and sequential circuits, arithmetic modules, and control logic. RTLLM v1.1 is widely adopted for functional correctness evaluation by previous works, while RTLLM v2.0 is used for systematic PPA assessment.

\paragraph{Evaluation Metrics.}
For functional correctness, we adopt the pass@k metric. For each task, the model generates $k$ candidate implementations and it is considered successfully solved if at least one candidate passes all provided testbenches.

For physical quality evaluation, we perform synthesis-based analysis using RTLLM v2.0 following ~\citep{chen2025chipseekr1generatinghumansurpassingrtl}. We report standard physical metrics including critical path delay, area, and power consumption. The evaluation pipeline uses Yosys and OpenROAD through NanGate45 process technology simulations.

\paragraph{Implementation Details.}
We conduct experiments using \textit{GPT-4o} and \textit{Gemini-3-Pro-Preview} as the backbone models and the sampling temperature is set to 0.8.
The Programmer Agent interacts with the Correctness Agent and the PPA Agent for at most two rounds respectively. Memory retrieval and triggering are performed by selecting the top-3 most relevant memory nodes from the memory bank based on semantic matching by LLM prompting. To reduce computational cost for high-frequency but easy operations, including memory retrieval and triggering, we employ a smaller model (\textit{GPT-4o-mini}). 

\begin{table*}[ht]
\centering
\small
\setlength{\tabcolsep}{4.5pt}
\renewcommand{\arraystretch}{1.15}

\begin{threeparttable}
\caption{
Functional comparison on VerilogEval and RTLLM v1.1.
Best results in each column are highlighted in bold.
}
\label{function_compare_new}

\begin{tabular}{l l rrr rrr rr}
\toprule
\textbf{Method} & \textbf{Base Model} &
\multicolumn{3}{c}{\textbf{VerilogEval-Machine (\%)}} &
\multicolumn{3}{c}{\textbf{VerilogEval-Human (\%)}} &
\multicolumn{2}{c}{\textbf{RTLLM v1.1 (\%)}} \\
\cmidrule(lr){3-5} \cmidrule(lr){6-8} \cmidrule(lr){9-10}
 & &
pass@1 & pass@5 & pass@10 &
pass@1 & pass@5 & pass@10 &
Syn@5 & Func@5 \\
\midrule
\multicolumn{10}{c}{\textit{Training-Based}} \\
\midrule
\multirow{2}{*}{RTLCoder}
 & Mistral 7B & 62.5 & 72.2 & 76.6 & 36.7 & 45.5 & 49.2 & 73.7 & 37.3 \\
 & DeepSeek-Coder 7B & 61.2 & 76.5 & 81.8 & 41.6 & 50.1 & 53.4 & 83.9 & 40.3 \\
\addlinespace

\multirow{3}{*}{BetterV}
 & CodeLlama 7B & 64.2 & 75.4 & 79.1 & 40.9 & 50.0 & 53.3 & -- & -- \\
 & DeepSeek-Coder 6.7B & 67.8 & 79.1 & 84.0 & 45.9 & 53.3 & 57.6 & -- & -- \\
 & CodeQwen 7B & 68.1 & 79.4 & 84.5 & 46.1 & 53.7 & 58.2 & -- & -- \\
\addlinespace

\multirow{3}{*}{CodeV}
 & CodeLlama 7B & 78.1 & 86.0 & 88.5 & 45.2 & 59.5 & 63.8 & 89.2 & 50.3 \\
 & DeepSeek-Coder 6.7B & 77.9 & 88.6 & 90.7 & 52.7 & 62.5 & 67.3 & 87.4 & 51.5 \\
 & CodeQwen 7B & 77.6 & 88.2 & 90.7 & 53.2 & 65.1 & 68.5 & 89.5 & 53.3 \\
\addlinespace

OriGen & DeepSeek-Coder 6.7B & 74.1 & 82.4 & 85.7 & 54.4 & 60.1 & 64.2 & -- & 65.5 \\
\addlinespace

\multirow{3}{*}{CraftRTL}
 & CodeLlama 7B & 78.1 & 85.5 & 87.8 & 63.1 & 67.8 & 69.7 & 93.9 & 52.9 \\
 & DeepSeek-Coder 6.7B & 77.8 & 85.5 & 88.1 & 65.4 & 70.0 & 72.1 & 92.9 & 58.8 \\
 & Starcoder2 15B & 81.9 & 86.9 & 88.1 & 68.0 & 72.4 & 74.6 & 93.9 & 65.8 \\
\addlinespace

ChipSeek-R1 & Qwen2.5-Coder 7B & 84.1 & 90.6 & 92.3 & 62.2 & 73.7 & 76.9 & 96.6 & 82.8 \\
\addlinespace

\midrule
\multicolumn{10}{c}{\textit{Training-Free}} \\
\midrule

\multirow{2}{*}{Direct}
 & GPT-4o & 65.9 & 71.4 & 72.7 & 57.1 & 63.9 & 66.7 & 93.9 & 65.5 \\
 & Gemini3-pro-preview & 88.8 & 92.3 & 92.3 & 91.7 & 94.2 & 94.9 & 100.0 & -- \\
\addlinespace

\multirow{2}{*}{VeriAgent}
 & GPT-4o & 86.7 & 90.2 & 90.2 & 78.2 & 85.9 & 89.1 & 100.0 & 72.4 \\
 & Gemini3-pro-preview & 97.9 & 99.3 & 99.3 & 98.1 & 98.7 & 98.7 & 100.0 & 89.7 \\

\bottomrule
\end{tabular}
\end{threeparttable}
\end{table*}

\subsection{Main Results}

\paragraph{Functional Correctness.}
Table~\ref{function_compare_new} reports functional correctness results on VerilogEval and RTLLM v1.1. 
Recent training-based approaches achieve strong performance by training small open-source models. In contrast, our framework directly leverages modern closed-source foundation models and achieves very high correctness. 
These observations are consistent with recent literature; e.g., ~\citet{zhao2025mage} reports 94.8\% pass@1 on VerilogEval-Human using closed-source models. 
Overall, these results suggest that functional correctness on current benchmarks is approaching saturation on existing benchmarks.

\begin{table*}[ht]

\centering
\small
\setlength{\tabcolsep}{7pt}
\renewcommand{\arraystretch}{1.15}
\newcommand{\ppa}[3]{#1\,/\,#2\,/\,#3}
\caption{
PPA comparison on RTLLM v2.0.
Each entry reports (Delay in ns / Area in $\mu m^2$ / Power in W).
N/A indicates no functionally correct and synthesizable design. Bold indicates the best PPA score.
}
\label{ppa_compare}

\resizebox{\linewidth}{!}{ 
\begin{tabular}{lcccc}
\toprule
\multirow{2}{*}{\textbf{Design}} 
& \textbf{Original} 
& \textbf{ChipSeek-R1} 
& \textbf{VeriAgent (gpt-4o)} 
& \textbf{VeriAgent (gemini3)} \\
& (ns / $\mu m^2$ / W) 
& (ns / $\mu m^2$ / W) 
& (ns / $\mu m^2$ / W) 
& (ns / $\mu m^2$ / W) \\
\midrule
accu & \ppa{0.47}{150.822}{1.23e-05} & \ppa{0.46}{210.672}{1.83e-05} & \ppa{0.53}{212.002}{1.78e-05} & \textbf{\ppa{0.49}{143.64}{1.14e-05}} \\
adder\_16bit & \ppa{0.84}{89.376}{6.49e-05} & \ppa{0.07}{93.632}{4.44e-05} & \textbf{\ppa{0.57}{97.888}{2.26e-06}} & \textbf{\ppa{0.57}{97.888}{2.26e-06}} \\
adder\_32bit & \ppa{0.76}{472.15}{0.000325} & \ppa{1.13}{191.786}{0.00012} & \ppa{1.08}{195.776}{4.52e-06} & \textbf{\ppa{0.51}{240.464}{5.6e-06}} \\
adder\_8bit & \ppa{0.35}{51.072}{3.14e-05} & \ppa{0.07}{46.816}{2.22e-05} & \textbf{\ppa{0.3}{49.21}{1.12e-06}} & \textbf{\ppa{0.3}{49.21}{1.12e-06}} \\
adder\_bcd & \ppa{0.34}{46.018}{3.85e-05} & \ppa{0.34}{46.018}{3.84e-05} & \textbf{\ppa{0.31}{42.56}{9.36e-07}} & \ppa{0.31}{42.826}{9.38e-07} \\
adder\_pipe\_64bit & \ppa{0.75}{2534.182}{0.000235} & N/A & \textbf{\ppa{0.1}{756.77}{4.43e-05}} & \ppa{0.14}{894.824}{4.99e-05} \\
alu & \ppa{1.92}{1573.39}{0.000751} & \ppa{1.75}{1286.908}{0.000433} & N/A & \textbf{\ppa{1.49}{1202.852}{2.73e-05}} \\
asyn\_fifo & \textbf{\ppa{0.72}{1397.032}{7.67e-05}} & N/A & N/A & N/A \\
barrel\_shifter & \ppa{0.17}{44.688}{1.41e-05} & \ppa{0.17}{39.368}{1.41e-05} & \ppa{0.17}{44.688}{8.62e-07} & \textbf{\ppa{0.17}{39.368}{8.13e-07}} \\
calendar & \ppa{0.44}{164.92}{1.44e-05} & \ppa{0.44}{164.92}{1.44e-05} & \ppa{0.54}{169.708}{1.4e-05} & \textbf{\ppa{0.4}{157.738}{1.3e-05}} \\
comparator\_3bit & \ppa{0.1}{11.704}{5.26e-06} & \ppa{0.11}{11.704}{5.25e-06} & \textbf{\ppa{0.1}{11.704}{3.03e-07}} & \textbf{\ppa{0.1}{11.704}{3.03e-07}} \\
comparator\_4bit & \ppa{0.16}{18.886}{8.91e-06} & \ppa{0.13}{17.29}{7.6e-06} & \ppa{0.13}{17.29}{4.33e-07} & \textbf{\ppa{0.12}{16.492}{3.99e-07}} \\
counter\_12 & \ppa{0.25}{36.176}{3.1e-06} & \ppa{0.25}{36.176}{3.1e-06} & \ppa{0.25}{36.176}{2.99e-06} & \textbf{\ppa{0.19}{31.92}{2.76e-06}} \\
div\_16bit & \ppa{5.18}{760.228}{0.027} & \ppa{5.57}{745.332}{0.0234} & \ppa{5.35}{755.706}{1.71e-05} & \textbf{\ppa{5.38}{725.382}{1.67e-05}} \\
edge\_detect & \ppa{0.12}{18.354}{1.79e-06} & \ppa{0.1}{17.822}{1.75e-06} & \ppa{0.12}{18.354}{1.73e-06} & \textbf{\ppa{0.13}{9.31}{7.64e-07}} \\
fixed\_point\_adder & \ppa{1.69}{606.214}{0.000565} & \textbf{\ppa{0.26}{24.294}{2.5e-05}} & \ppa{0.66}{220.248}{5.09e-06} & \ppa{0.66}{220.248}{5.09e-06} \\
fixed\_point\_subtractor & \ppa{1.09}{477.736}{0.000381} & \textbf{\ppa{0.15}{20.482}{9.44e-06}} & \ppa{0.56}{95.228}{2.18e-06} & \ppa{0.56}{95.228}{2.18e-06} \\
freq\_divbyeven & \ppa{0.25}{40.166}{3.63e-06} & N/A & N/A & \textbf{\ppa{0.17}{20.482}{1.87e-06}} \\
freq\_divbyfrac & \ppa{0.2}{48.678}{4.6e-06} & N/A & N/A & \textbf{\ppa{0.15}{32.452}{3.14e-06}} \\
freq\_divbyodd & \ppa{5.17}{59.052}{6.63e-06} & \ppa{5.18}{62.776}{6.91e-06} & N/A & \textbf{\ppa{0.1}{36.708}{3.61e-06}} \\
instr\_reg & \ppa{0.14}{117.04}{1.03e-05} & \ppa{0.14}{117.306}{1.03e-05} & \textbf{\ppa{0.14}{114.912}{9.7e-06}} & \textbf{\ppa{0.14}{114.912}{9.7e-06}} \\
JC\_counter & \ppa{0.1}{340.48}{3.54e-05} & \ppa{0.1}{340.48}{3.54e-05} & \textbf{\ppa{0.1}{340.48}{3.41e-05}} & \textbf{\ppa{0.1}{340.48}{3.41e-05}} \\
LFSR & \textbf{\ppa{0.14}{25.004}{2.45e-06}} & \textbf{\ppa{0.14}{25.004}{2.45e-06}} & N/A & \ppa{0.15}{24.206}{2.51e-06} \\
LIFObuffer & \ppa{0.38}{226.1}{0.00027} & \ppa{0.36}{216.79}{0.000137} & \ppa{0.38}{216.79}{4.05e-06} & \textbf{\ppa{0.28}{213.066}{4.23e-06}} \\
multi\_16bit & \ppa{2.03}{933.394}{7.36e-05} & \ppa{1.97}{935.522}{7.36e-05} & \ppa{2.14}{881.79}{7.04e-05} & \textbf{\ppa{0.79}{553.28}{4.48e-05}} \\
multi\_8bit & \ppa{1.5}{483.854}{0.00085} & \ppa{0.79}{373.996}{5.45e-05} & \ppa{0.75}{328.776}{7.58e-06} & \textbf{\ppa{0.75}{327.978}{7.55e-06}} \\
multi\_pipe\_4bit & \ppa{0.34}{174.762}{1.51e-05} & \textbf{\ppa{0.11}{54.546}{1.28e-05}} & \ppa{0.1}{154.546}{1.01e-05} & \ppa{0.1}{154.546}{1.01e-05} \\
multi\_pipe\_8bit & \ppa{0.8}{874.608}{7.52e-05} & N/A & \ppa{0.78}{686.28}{5.45e-05} & \textbf{\ppa{0.69}{717.668}{5.59e-05}} \\
parallel2serial & \ppa{0.2}{48.678}{4.5e-06} & \ppa{0.19}{47.082}{4.36e-06} & N/A & \textbf{\ppa{0.17}{45.486}{4.14e-06}} \\
pe & \ppa{1.27}{3651.382}{0.000224} & \ppa{1.27}{3651.382}{0.000224} & \textbf{\ppa{1.22}{3685.696}{0.000104}} & \ppa{1.27}{3649.786}{0.000102} \\
pulse\_detect & \ppa{0.18}{17.556}{1.52e-06} & \textbf{\ppa{0.17}{16.226}{1.47e-06}} & N/A & N/A \\
radix2\_div & \textbf{\ppa{0.59}{414.162}{3.39e-05}} & N/A & N/A & N/A \\
RAM & \ppa{0.25}{635.74}{5.56e-05} & \textbf{\ppa{0.19}{475.076}{3.92e-05}} & \ppa{0.21}{479.864}{3.68e-05} & \ppa{0.21}{476.938}{3.65e-05} \\
right\_shifter & \textbf{\ppa{0.08}{36.176}{4.32e-06}} & \textbf{\ppa{0.08}{36.176}{4.32e-06}} & \textbf{\ppa{0.08}{36.176}{4.32e-06}} & \textbf{\ppa{0.08}{36.176}{4.32e-06}} \\
ring\_counter & \ppa{0.1}{46.816}{4.7e-06} & \ppa{0.1}{40.964}{4.01e-06} & \ppa{0.18}{32.452}{2.54e-06} & \textbf{\ppa{0.1}{31.92}{2.67e-06}} \\
ROM & \ppa{0.14}{6.65}{8.85e-07} & \ppa{0.14}{6.65}{8.85e-07} & \textbf{\ppa{0.14}{6.65}{1.48e-07}} & \textbf{\ppa{0.14}{6.65}{1.48e-07}} \\
sequence\_detector & \ppa{0.15}{36.442}{3.35e-06} & \ppa{0.19}{25.277}{2.27e-06} & N/A & \textbf{\ppa{0.1}{22.61}{2.22e-06}} \\
serial2parallel & \ppa{0.4}{156.142}{1.4e-05} & \ppa{0.28}{157.738}{1.39e-05} & N/A & \textbf{\ppa{0.28}{157.206}{1.34e-05}} \\
signal\_generator & \ppa{0.37}{93.1}{7.54e-06} & \ppa{0.38}{74.214}{6.08e-06} & \ppa{0.3}{78.47}{6.2e-06} & \textbf{\ppa{0.24}{52.402}{4.43e-06}} \\
square\_wave & \ppa{0.41}{100.282}{8.39e-06} & \ppa{0.41}{100.282}{8.39e-06} & \textbf{\ppa{0.33}{81.396}{7.27e-06}} & \textbf{\ppa{0.33}{81.396}{7.27e-06}} \\
sub\_64bit & \ppa{2.3}{404.586}{0.000268} & \ppa{2.08}{400.862}{0.000271} & \ppa{1.7}{454.328}{1.06e-05} & \textbf{\ppa{2.13}{405.118}{9.25e-06}} \\
traffic\_light & \ppa{0.37}{149.758}{1.31e-05} & \ppa{0.39}{139.916}{1.22e-05} & N/A & \textbf{\ppa{0.32}{90.174}{7.59e-06}} \\
up\_down\_counter & \ppa{0.7}{217.854}{1.74e-05} & \ppa{0.67}{188.86}{1.61e-05} & \ppa{0.53}{183.54}{1.41e-05} & \textbf{\ppa{0.41}{161.994}{1.29e-05}} \\
width\_8to16 & \ppa{0.24}{186.732}{1.67e-05} & \textbf{\ppa{0.21}{173.698}{1.62e-05}} & \ppa{0.24}{186.732}{1.62e-05} & \ppa{0.24}{180.348}{1.63e-05} \\

% Geo Mean
\midrule 
\textbf{Geometric Mean}  & \ppa{0.417}{142}{2.65e-05} & \ppa{0.329}{112}{1.93e-05} & \ppa{0.362}{122}{6.89e-06} & \ppa{0.294}{109}{5.72e-06} \\

% Relative PPA Score
\textbf{Relative PPA Score}  & 1.000 & 2.196 & 5.119 & 8.539 \\
\bottomrule

\end{tabular}
} 
\end{table*}

\paragraph{PPA Scores}

We further evaluate the physical quality of generated RTL implementations using RTLLM v2.0. 
Following ~\citet{chen2025chipseekr1generatinghumansurpassingrtl}, Table~\ref{ppa_compare} reports PPA results obtained using the NanGate45 technology library. 
Following~\citet{chen2025chipseekr1generatinghumansurpassingrtl}, we adopt a composite metric termed \textit{Relative PPA Score} to summarize overall hardware efficiency. 
The PPA score is defined as

\begin{equation}
\text{PPA Score} = \frac{1}{\text{Delay} \times \text{Area} \times \text{Power}},
\end{equation}

where higher values indicate better combined hardware efficiency. 
The relative PPA score is obtained by normalizing each design with respect to the original reference implementation provided by the benchmark. 
For transparency, Table~\ref{ppa_compare} also reports the raw delay, area, and power values for each design. Overall, VeriAgent achieves the best results on most individual metrics as well as the highest mean relative PPA score.

\begin{figure}[htbp]
\centering
\includegraphics[width=1.0\linewidth]{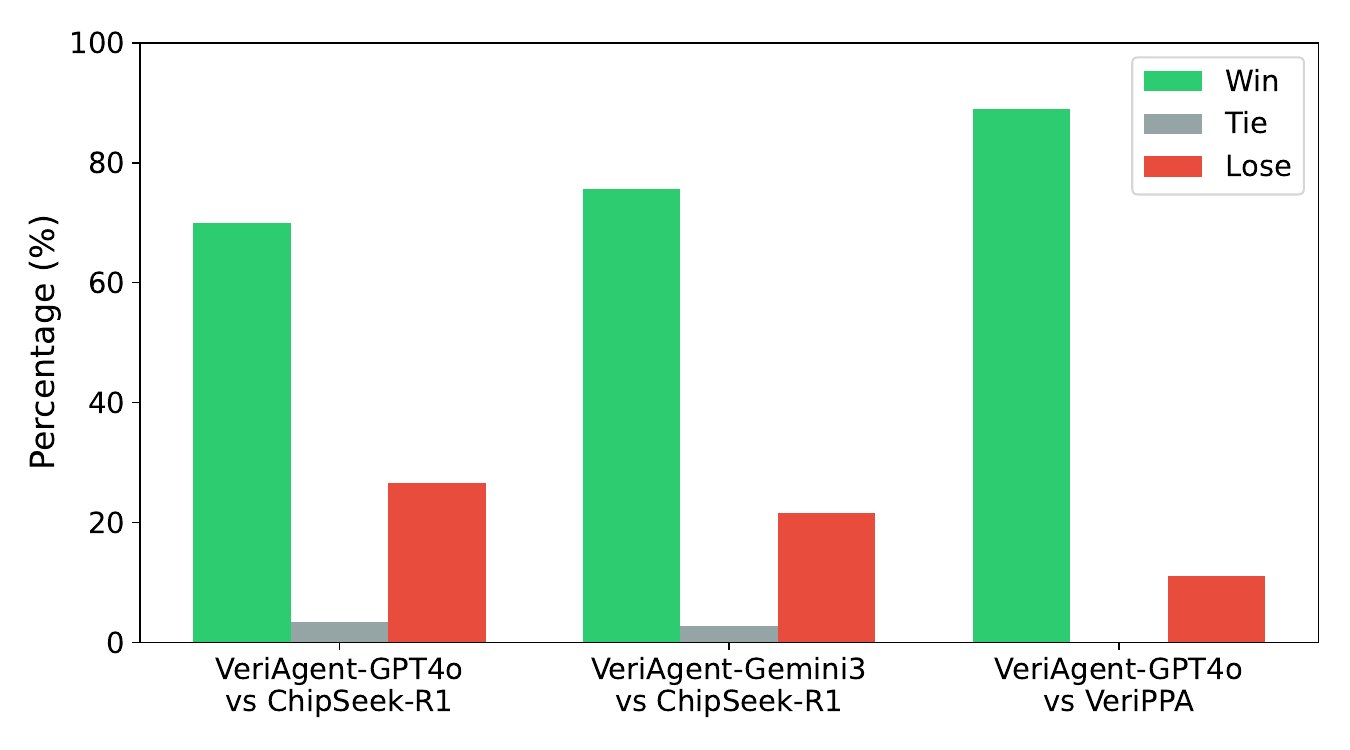}
\caption{Win rate comparison between VeriAgent and prior methods.}
\label{fig:winrate}
\end{figure}
\paragraph{PPA Win Rate}
To further understand the effectiveness of our optimization framework, we perform a pairwise comparison between VeriAgent and prior optimization systems. Both VeriPPA~\citep{thorat2025llmveri} and VeriOpt~\citep{tasnia2025veriopt} adopt in-context learning~\citep{wu2023openicl} for PPA  optimization. Since VeriOpt uses a different evaluation protocol and dataset from ours and does not release detailed implementation information, we compare the results of VeriPPA reported in its original paper for comparison. 
Specifically, we compute the \textit{win rate} between two methods by comparing their PPA scores on tasks where both methods successfully produce synthesizable designs. A method is considered to win on a task if its PPA score is higher.

Figure~\ref{fig:winrate} illustrates the win rate comparison between VeriAgent and two representative baselines: VeriPPA and ChipSeek-R1.
It is worth noting that VeriPPA~\citep{thorat2025llmveri} successfully optimizes only a small subset of tasks, while our framework consistently produces valid implementations for a significantly larger portion of the benchmark.
Across the overlapping tasks, VeriAgent demonstrates a higher win rate against both baselines, indicating stronger optimization capability in discovering hardware-efficient implementations.

\subsection{Ablation Study}

\begin{table}[t]
\centering
\caption{Ablation study of VeriAgent. Relative PPA Score (higher is better).}
\label{table:ablation}
\begin{tabular}{lcc}
\toprule
\textbf{Method} & \textbf{GPT-4o} & \textbf{Gemini-3} \\
\midrule
\textbf{VeriAgent (Full)} & 5.119 & 8.539 \\
\quad w/o Memory & 3.839 & 6.541 \\
\quad w/o Tool Feedback & 3.344 & 5.189 \\
\quad w/o PPA Analyzer & 2.724 & 3.405 \\
\bottomrule
\end{tabular}
\end{table}

To understand the contribution of individual components in our framework, we conduct an ablation study by progressively removing key modules from VeriAgent. The evaluated variants include:

\begin{itemize}
\item \textbf{w/o Memory}: the memory manager is disabled. The agents generates RTL code solely based on the current task context without historical optimization guidance.
\item \textbf{w/o Tool Feedback}: disabling synthesis and simulation feedback from EDA tools. The PPA Agent remains active but performs PPA-oriented reasoning without access to tool-generated signals.
\item \textbf{w/o PPA Agent}: removing the PPA Agent.
\end{itemize}

All ablation variants use the same base LLM, prompt templates, interaction rounds, and evaluation protocol as the full VeriAgent system. Table~\ref{table:ablation} presents the relative PPA scores achieved by different variants. The ablation variants are constructed progressively, where each configuration removes one additional component based on the previous variant.

Removing the memory module leads to a noticeable performance drop, indicating that accumulated design experience plays an important role in guiding optimization. While an ideal system could rely on a manually curated memory bank designed by hardware experts, our memory evolution mechanism automatically constructs and refines such knowledge during execution, enabling the agents to reuse effective optimization strategies.
Disabling tool feedback further degrades performance, highlighting the importance of synthesis and simulation signals in guiding the search process. Without these signals, the agents lack reliable feedback to evaluate candidate implementations and identify optimization opportunities.
These results demonstrate that each component contributes to the overall effectiveness of VeriAgent.

%% file: section/2-related.tex
\section{Related Works}

\subsection{RTL Code Generation}

Automated RTL code generation~\citep{zhang2016automatic} has long been a key objective in electronic design automation. Traditional approaches~\citep{takamaeda2015pyverilog, amjad2019verilog} typically rely on rule-based synthesis from high-level specifications or domain-specific hardware description languages~\citep{palnitkar2003verilog, ashenden2007digital}. While effective in well-structured settings, these techniques often lack flexibility when dealing with diverse natural-language specifications. 

With the emergence of large language models, researchers have explored direct RTL code generation from natural language descriptions. 
% Recent benchmarks such as VerilogEval~\citep{liu2023verilogeval} and RTLLM~\citep{lu2024rtllm} demonstrate that modern LLMs can generate syntactically correct and functionally valid Verilog implementations for a wide range of hardware modules. 
Existing approaches for LLM-based RTL code generation often adapt LLMs to hardware description tasks through supervised fine-tuning~\citep{liu2024rtlcoder,pei2024betterv,zhao2025codev,cui2024origen} or reinforcement learning~\citep{chen2025chipseekr1generatinghumansurpassingrtl,zhang2025qimeng,li2025autotriton} on large-scale realistic and synthetic datasets~\citep{wang2025epicoder}. Although such methods improve task-specific performance, the generated datasets often exhibit strong distribution biases, limiting generalization to complex real-world designs. Training-free approaches~\citep{ho2025verilogcoder,zhao2025mage} instead leverage prompting strategies, iterative feedback and integration with commercial LLMs without modifying model parameters.  More recently, a few works have begun to consider physical design objectives: VeriOpt~\citep{tasnia2025veriopt} and VeriPPA~\citep{thorat2025llmveri} integrate PPA-related signals from synthesis tools into the generation process. However, these approaches typically treat tool feedback as one-shot information and lack mechanisms to retain and reuse optimization knowledge across multiple iterations. In contrast, our framework explicitly externalizes both structural patterns and EDA-derived signals into a reusable structured memory, enabling continuous, feedback-driven optimization that jointly improves PPA metrics.

\subsection{Multi-Agent Systems for LLMs}

Recent advances in Large Language Models have led to increasing interest in multi-agent systems that coordinate multiple LLM-based agents to solve complex tasks~\citep{wang2025tdag}. Instead of relying on a single model invocation, these frameworks decompose problems into specialized roles such as planning, generation, critique, and verification, enabling collaborative reasoning and iterative refinement.
Several works explore role-based collaboration among LLM agents. ~\citet{Chatdev} propose ChatDev, a multi-agent framework that encodes software engineering workflows as standardized operating procedures, allowing agents with different roles to cooperatively complete complex tasks. Similarly, AutoGen~\citep{wu2023autogenenablingnextgenllm} provides a general framework for building LLM applications through multi-agent conversations, supporting flexible interactions between agents, humans, and external tools. These systems demonstrate that structured collaboration can significantly improve reliability and task decomposition.
Another line of work focuses on improving reasoning quality through agent interaction. Reflexion~\citep{shinn2023reflexion} introduces a self-reflection mechanism where an agent iteratively improves its outputs based on verbal feedback generated from previous attempts. Multi-Agent Debate~\citep{liang2024encouraging} extends this idea by allowing multiple agents to argue and critique each other before producing a final solution, encouraging diverse reasoning paths.